\documentclass[11pt,a4paper]{article}
\usepackage[hyperref]{emnlp2020}
\usepackage{times}
\usepackage{latexsym}
\usepackage{booktabs}
\usepackage{graphicx}

\usepackage{microtype}

\aclfinalcopy % Uncomment this line for the final submission
\title{Unsupervised Parallel Corpus Mining on Web Data}

\author{Guokun Lai \\
  Carnegie Mellon University \\
  \texttt{guokun@cs.cmu.edu} \\\And
  Zihang Dai \\
  Carnegie Mellon University \\
  \texttt{dzihang@cs.cmu.edu} \\\And
  Yiming Yang \\
  Carnegie Mellon University \\
  \texttt{yiming@cs.cmu.edu}}

\date{}

\usepackage{amsmath,amsfonts,bm}

\DeclareMathAlphabet{\mathsfit}{\encodingdefault}{\sfdefault}{m}{sl}
\SetMathAlphabet{\mathsfit}{bold}{\encodingdefault}{\sfdefault}{bx}{n}

\def\gA{{\mathcal{A}}}
\def\gB{{\mathcal{B}}}
\def\gC{{\mathcal{C}}}

\def\gF{{\mathcal{F}}}

\def\gM{{\mathcal{M}}}

\makeatletter
\def\rbr{\@ifnextchar[{\roundbracket}{\roundbracket*}}
\def\sbr{\@ifnextchar[{\squarebracket}{\squarebracket*}}
\def\cbr{\@ifnextchar[{\curlybracket}{\curlybracket*}}
\makeatother

\begin{document}

\maketitle
	
\begin{abstract}
With a large amount of parallel data, neural machine translation systems are able to deliver human-level performance for sentence-level translation. 
However, it is costly to label a large amount of parallel data by humans. 
In contrast, there is a large-scale of parallel corpus created by humans on the Internet. 
The major difficulty to utilize them is how to filter them out from the noise website environments. Current parallel data mining methods all require labeled parallel data as the training source.   
In this paper, we present a pipeline to mine the parallel corpus from the Internet in an unsupervised manner. 
On the widely used WMT'14 English-French and WMT'16 English-German benchmarks, the machine translator trained with the data extracted by our pipeline achieves very close performance to the supervised results.
On the WMT'16 English-Romanian and Romanian-English benchmarks, our system produces new state-of-the-art results, 39.81 and 38.95 BLEU scores, even compared with supervised approaches.   
\end{abstract}

\section{Introduction}
% Machine translation is one of the most successful machine learning applications in the field of natural language processing \cite{vaswani2017attention,ott2018scaling}. 
% Especially when the targeted language pair has large enough parallel corpus, the machine translation system is able to produce the results similar to humans. 
% But most of the parallel corpus is labeled on the language pair of English and another language. 
% In order to translate the sentences between arbitrary language pair $(a, b)$, we usually use the English as the pivot language, $a\rightarrow \mbox{En} \rightarrow b$. 
% However, the two-step process of the pivoted method may introduce the additional error, which leads to lower translation accuracy compared with the directly translation method.

%On the other hand, the unsupervised machine translation technique has evolved rapidly recently \cite{lample2017unsupervised,artetxe2017unsupervised,lample2018phrase}. 
%With the help of a multi-lingual pretrained model \cite{devlin2018bert}, the unsupervised machine translation system has achieved a reasonable performance \cite{lample2019cross,song2019mass}. 
%Although there is still an obvious gap between their performance and the supervised model results, it is able to generate pseudo parallel samples, which potentially could be used to teach a machine learning model to differentiate the parallel and un-parallel samples. 

As one the the most successful applications in natural language processing~\cite{vaswani2017attention,ott2018scaling}, modern neural machine translation systems are able to match human-level performances given a large amount of labeled parallel data.
Despite the success, it remains extremely challenging to construct a large parallel corpus for a new language pair given the non-trivial skill requirement and annotation cost.
% In fact, most of the existing parallel corpora are between English and other languages.
% Hence, in order to translate sentences between a non-English language pair $(a, b)$, a common practice is to use English as the pivot language and utilize two separately trained translation systems $a \rightarrow \mbox{En} \rightarrow b$. 
% However, the two-step process of the pivoted method often suffers from compounding errors.

% Machine translation is one of the most successful machine learning applications in the field of natural language processing \cite{vaswani2017attention,ott2018scaling}. 
% Especially when the targeted language pair has large enough parallel corpus, the machine translation system is able to produce the results similar to humans. 
% But most of the parallel corpus is labeled on the language pair of English and another language. 
% In order to translate the sentences between arbitrary language pair $(a, b)$, we usually use the English as the pivot language, $a\rightarrow \mbox{En} \rightarrow b$. 
% However, the two-step process of the pivoted method may introduce the additional error, which leads to lower translation accuracy compared with the directly translation method.

On the other hand, there exists a large quantity of unaligned sentences expressing the same or very similar meanings in different languages.
For any language pair, if we can correctly extract and pair such sentences with similar meanings in corresponding languages, they could be used as \textit{crawled parallel corpus} to train the machine translation system directly.
In fact, this idea has taken by the parallel corpus mining community, which has led to various pseudo parallel corpus between European languages\footnote{https://paracrawl.eu/index.php/news/item/9-paracrawl-works} and hence improved performances~\cite{sanchez2018prompsit,azpeitia2018extracting,artetxe2018margin}.
Despite the success, methods along this line still require a significant amount of \textit{labeled parallel corpus} to train a sentence aligner, which is then used to filter the abundant unaligned text.
This requirement restricts the practical application of these parallel corpus mining methods.

% On the other hand, a large quantity of parallel data created by human exist on the Internet. For example, the ParaCrawl \cite{espla2019paracrawl}, a web parallel data mining project, crawls a set of the parallel corpus for the European languages. 
% By utilizing web parallel data, they achieve better BELU scores compared with the models trained with WMT supervision data\footnote{https://paracrawl.eu/index.php/news/item/9-paracrawl-works}.
% In order to extract the parallel data from the noise website environment, many parallel corpus mining algorithms have been developed \cite{sanchez2018prompsit, azpeitia2018extracting, artetxe2018margin}. 
% All these methods are conditioned on a labeled parallel corpus as the training data, which hinders us to apply this method to the language pair without the supervision data. 
In the meantime, the unsupervised machine translation technique has developed rapidly.
It provides us a potential choice to generate the pseudo parallel data based on the unsupervised machine translator, and use them as the training data for the parallel data miner. 

Based on this intuition, we propose an unsupervised web parallel corpus mining pipeline by combining the unsupervised machine translation with the web parallel corpus mining technique.
It can automatically collect and extract the high-quality parallel data from the Internet without requiring any labeled data. 
The propose pipeline reduces the cost of collecting the parallel data for arbitrary language pairs.
In our experiment, we show that the machine translation system trained with crawled parallel data from our system is able to achieve a similar or even superior performance compared to fully supervised systems on the WMT benchmarks. 

Our proposed pipeline can be separated into three phases:
(1) Train an unsupervised machine translation model and use it to generate pseudo parallel corpus $\gA$. 
(2) Construct a dictionary based on the pseudo parallel data. The crawler will collect the raw parallel corpus $\gB$ from the Internet based on the generated dictionary. 
(3) Use the pseudo parallel data $\gA$ to train a classifier to differentiate whether a pair of sentences is a parallel sample of the given language pair.
We use the classifier to filter $\gB$ to get the final parallel corpus $\gC$.
Finally, we treat $\gC$ as supervised data to train the machine translation system. 
The details of the pipeline are described in section \ref{sec:model} and the experiment results are included in section \ref{sec:exp}. 
\section{The Proposed Pipeline}
\label{sec:model}

In this section, we introduce the details of the proposed unsupervised web parallel corpus mining pipeline. In the following parts, the targeted language pair is denoted as $(p,q)$.

\paragraph{Train an Unsupervised Machine Translation System:}
In the first step, we follow the XLM paper \cite{lample2019cross} to train an unsupervised machine translator, denote as $\gF$. The training process is to initialize the encoder and decoder by the pretrained XLM model, 
then minimize the objective function which combines the de-noising encoder-decoder loss and the back-translation loss. 
Next, given the monolingual data of language $p$, $\gM_p$, we can generate a pseudo parallel corpus $\gA_{(p,q)} = (\gM_p, \gF(\gM_p))$.

\paragraph{Obtain a Dictionary:} 
To run the mining crawler, we need a dictionary for language pair $(p,q)$ as the seed. Here, we run a statistical machine translation model \cite{koehn2007factored} on the pseudo parallel corpus $\gA_{(p,q)}$ to generate a dictionary. 

\paragraph{Crawl the Parallel Data:} To crawl the parallel data from Internet, we utilize Bitextor package\footnote{https://github.com/bitextor/bitextor} \cite{espla2009bitextor} as our crawler. Given a website URL, the crawler would download all HTML pages from its domain. Then the package performs two-stage processing, document and sentence alignments, to generate aligned sentence pairs. 

In the document alignment step, the algorithm will take the URL and HTML structure information of pages as input to align website pages. For example, the pages with URLs, ``xx.com/abc/en'' and ``xx.com/abc/de'', would produce high probability to be aligned. 

After aligning documents, the algorithm utilize the Hunalign 
\cite{varga2007parallel} package to align the sentences in the paired documents. It takes the dictionary, generated in the last step, and linguistic information of sentences as the input, and produces the aligned sentence pairs.

The Bitextor package allows users to integrate machine learning system into the document and sentence alignment process, which can improve the precision. Here, we could inject the machine translator trained in the first step. But in practical, we found that the neural machine translator would be the speed bottleneck of the crawling pipeline. So we did not use this function of Bitextor.

\paragraph{Filter the Crawled Data:}
The first step of filtration is following the heuristic rules described in \cite{artetxe2018margin}. It includes three rules: (1) remove all duplicate sample pairs. (2) remove any sentences whose length small than 4. (3) remove any sample pairs whose overlap ratio is larger than 50\%. After applying these rules, nearly 80\% crawled parallel data are removed. Toward this point, we denote the outcoming parallel corpus as $\gB_{(p,q)}$.
 
Because in the previous parts of proposed pipeline, we only use a learned dictionary to mine the parallel corpus, which limits the precision of crawler. Simultaneously, in order to keep most of high-quality parallel data, we set a low alignment threshold to promise a high recall rate.

Next, we need to perform a post-process to filter out the high quality parallel sentence from the noise data $\gB_{(p,q)}$. We use the pseudo parallel data generated in the first step $\gA_{(p,q)}$ to train a classifier to differentiate the parallel and unparallel sentence pairs. We treat $\gA_{(p,q)}$ as the positive samples, and randomly generate negative samples by sampling the unpaired sentences from $\gA_{(p,q)}$. Here, we train two machine learning classifiers:
\begin{itemize}
	\item Random Forest: We use the Bicleaner \cite{sanchez2018prompsit} tool\footnote{https://github.com/bitextor/bicleaner} to train a random forest classifier. This classifier can perform fast inference on CPU. So it can be integrated into the crawler step to save the disk memory for the intermediate results.
	\item Finetuned XLM: We finetune a XLM model as another classifier, which is the state-of-the-art method for the text classification. Due to its computation cost for inference, we uses this classifier after collecting the results from crawler step. 
\end{itemize}

After two-step filtration, we obtain a high-quality parallel dataset $\gC_{(p,q)}$. We can use it with any supervised machine translation algorithm to train the final machine translator.

\section{Experiment}
\label{sec:exp}

\subsection{Experiment settings}

In this section, we will test the proposed pipeline on three language pairs, English-French, English-German, and English-Romanian.
In the first step to generate pseudo parallel corpus $\gA_{(p,q)}$, we follow the training script in the XLM repository\footnote{https://github.com/facebookresearch/XLM} to train the unsupervised machine translator. Next, we sample 1M sentence from NewCrawl\footnote{https://www.statmt.org/wmt16/translation-task.html} datasets of French, German and Romanian, and translate them into English by the unsupervised machine translator $\gF$ to obtain $\gA_{(p,q)}$. 

%After obtaining the dictionary from $\gA_{(p,q)}$, we filter out the English word that did not appear in a pre-defined English vocabulary, whose size is 40k. Then we got an English-French, English-German and English-Romanian dictionary, whose size is 19k, 25k and 20k respectively. 

For the URL domains feed into the crawler, we follow the ones used in the ParaCrawl project \cite{espla2019paracrawl}, whose statistic information is included in the table \ref{tab:url} . 

For the finetuned XLM model in the filtration step, we use the pretrained 6-layer XLMs, which are the same ones in the first step, as the initial parameters,  then finetune them on the $\gA_{(p,q)}$ for 10 epochs. The hyperparameter setting is the same as the XNLI finetuning script in the XLM repository.

\begin{table}[!ht]
	\centering
	\begin{tabular}{l|ccc}
		\toprule
		Language Pair  & En-Fr & En-De & En-Ro \\
		\midrule
		\# url domains & 62.5K & 84.5K & 12.8K \\
		\bottomrule
	\end{tabular}
	\caption{The sizes of crawled URL domains}
	\label{tab:url}
    \vspace{-0.5cm}
\end{table}

\subsection{The Results of Crawling Pipeline}

In the table \ref{tab:pip-result}, we summarize the result of unsupervised web parallel data mining pipeline.
Firstly, we observe that the size of crawled data has a similar scale of supervision data in WMT benchmark. 
Here, the WMT of EN-Fr indicates WMT2014 training set, and WMT of En-De and En-Ro are WMT2016 training set. 
Secondly, The result of the filtration process, comparing the size of $\gB_{(p,q)}$ and $\gC_{(p,q)}$,  indicates 40\%-50\% crawled data are not high-quality parallel data. 

In the following parts, we are going to evaluate the quality of this parallel corpus $\gC_{(p,q)}$ by using it to train neural machine learning systems and compare the system performance on the supervised and unsupervised machine translation benchmark results. 

\begin{table}[!ht]
	\centering
	\begin{tabular}{l|cc|c}
		\toprule
		parallel set & $|\gB_{(p,q)}|$ & $|\gC_{(p,q)}|$ & WMT   \\
		\midrule
		En-Fr        & 21.2M           & 12.0M           & 35.7M \\
		En-De        & 22.6M           & 10.6M           & 3.96M \\
		En-Ro        & 1.23M           & 724K            & 399K  \\
		\bottomrule
	\end{tabular}
	\caption{The sizes of the crawled and filtered parallel corpus}
	\label{tab:pip-result}
	\vspace{-0.5cm}
\end{table}

%\begin{table*}[!ht]
%	\centering
%	\begin{tabular}{l|cccccc}
%		\toprule
%		Data                          & En-Fr & Fr-En & En-De & De-En & En-Ro & Ro-En \\
%		\midrule
%		XLM+Crawled Data              & 38.81 & 38.00 & 32.92 & 41.46 & 39.69 & 38.95 \\
%		XLM+Crawled Data wo Filter    & 37.23 & 37.87 & 32.23 & 40.77 & 39.00 & 38.27 \\
%		%Random+Crawled Data wo Filter & 36.38 & 35.57 & 30.62 & 37.64 & 36.22 & 35.19 \\
%		\bottomrule
%	\end{tabular}
%	\caption{The ablation study of the filtration process. The experiment setting is identical to the unsupervised machine translation experiment in table \ref{tab:usmt}.}
%	\label{tab:nofilter-usmt}
%\end{table*}

\subsection{Evaluation with Supervised Machine Translation Benchmarks}
\label{sec:sup}
Firstly, we evaluate the parallel corpus $\gC_{(p,q)}$ with the supervised machine translation benchmark. 
We follow the experiment setting in the Scaling NMT paper \cite{ott2018scaling}, including model architecture and choice of the hyper-parameters, and report the BELU score on the En-Fr and En-De directions on the WMT2014 test sets.

The evaluation results are included in table \ref{tab:smt}. WMT indicates that the model trained with the WMT training set. 
bt means the back-translation augmentation. 
From the results, we obverse that the machine translation system trained with $\gC_{(p,q)}$ can achieve similar performance to the ones trained with millions of human-labeled parallel samples. 
The performance gap is small than 1 BELU score
It indicates that the quality of $\gC_{(p,q)}$ is similar to the current largest-scale public parallel dataset, while the proposed website data mining pipeline does not require any labeled parallel sample and dictionary as the seed. 

\begin{table}[!ht]
	\centering
	\resizebox{\linewidth}{!}
	{
		\begin{tabular}{l|cc}
			\toprule
			Data                                 & En-Fr & En-De \\
			\midrule
			WMT\cite{ott2018scaling}             & 43.2  & 29.3  \\
			%WMT+pc\cite{artetxe2018margin}       & -     & 31.8  \\
			WMT+bt\cite{edunov2018understanding} & 45.6  & 35.0  \\
			\midrule
			Crawled Data                         & 42.79 & 28.66 \\
			\bottomrule
		\end{tabular}
	}
	\caption{Evaluation of the crawled corpus $\gC_{(p,q)}$ on the supervised machine translation benchmark}
	\label{tab:smt}
    \vspace{-0.5cm}
\end{table}

\begin{table*}[!ht]
	\centering
	\begin{tabular}{l|cccccc}
		\toprule
		Model                           & En-Fr & Fr-En & En-De & De-En & En-Ro & Ro-En \\
		\midrule
		XLM\cite{lample2019cross}       & 33.4  & 33.3  & 27.0  & 34.3  & 33.3  & 31.8  \\
		MASS\cite{song2019mass}         & 37.5  & 34.9  & 28.3  & 35.2  & 35.2  & 33.1  \\
		mBart\cite{liu2020multilingual} & -     & -     & 29.8  & 34.0  & 35.0  & 30.5  \\
		\midrule
		Crawled Data+XLM                & 38.81 & 38.00 & \textbf{32.92} & \textbf{41.46} & \textbf{39.96} & \textbf{38.95} \\
		Crawled Data+Mass               & \textbf{39.61} & \textbf{38.65} & 32.85 & 40.76 & 39.81 & 38.91 \\
		\bottomrule
	\end{tabular}
	\caption{Evaluation of the crawled corpus $\gC_{(p,q)}$ on the unsupervised machine translation benchmark setting. }
	\label{tab:usmt}
    \vspace{-0.5cm}
\end{table*}

\subsection{Evaluation with Unsupervised Machine Translation Benchmarks}

Next, we evaluate our corpus on the benchmark setting of unsupervised machine translation problems. Similar to its problem definition, our pipeline can train a machine translation system without requiring any labeled parallel samples. 
The model architecture design and choice of the hyper-parameters are the same as XLM \cite{lample2019cross} and MASS \cite{song2019mass} papers. 
The machine translation systems are trained with $\gC_{(p,q)}$ and the back-translation augmented data generated in an online manner. 
The (En, Fr) results are the BLEU scores on the WMT2014 test set. 
The (En, De) and (En, Ro) results are the BELU scores on the WMT2016 test set. 

The experiment results are included in table \ref{tab:usmt}. Compared with both baselines, the model trained with data from the proposed pipeline achieves a large margin improvement in all directions. The proposed method averagely improves 4.55 BELU scores compared with the best baseline. In the low resource case, Ro-En, our result, 38.95 BELU score, achieves new state-of-the-art results, even compared with the best performance with the WMT supervision data, which is 38.5 BELU score.

\begin{table}[!ht]
	\centering
	\begin{tabular}{l|cc}
		\toprule
		Supervision Data                   & En-Fr & En-De \\
		\midrule
		$\gC_{(p,q)}$           & 42.79 & 28.66 \\
		$\gB_{(p,q)}$           & 42.24 & 28.02 \\
		$\gB_{(p,q)} - \gC_{(p,q)}$   & 19.71 & 24.91 \\
		\bottomrule
	\end{tabular}
	\caption{Ablation study of the filtration process}
	\label{tab:nofilter-smt}
\end{table}

\subsection{Ablation Study about the Post Filtration}

To better understand the importance of the crawler and filtration components, we perform an ablation study by eliminating the parallel data classifier in the filtration process from the proposed pipeline. 
We train three models respectively with the filtered parallel data $\gC_{(p,q)}$, raw parallel data $\gB_{(p,q)}$, and the low quality data $\gB_{(p,q)} - \gC_{(p,q)}$, which are the samples discarded by the classifier.  
The experiment setting is same as the supervised machine translation study in section \ref{sec:sup}.
The experiment results are present in table \ref{tab:nofilter-smt}. Surprisingly, trained with raw parallel data, the model can achieve similar performance compared to the filtered version, where the difference is smaller than 1 BELU score. 
On the other hand, the models trained with low-quality parallel data have significantly lower performance.  
It indicates that the filtration process can differentiate the quality of the parallel samples, but leaving this noise in the neural machine translator training process would not harm the final performance too much.

\section{Related Work}

The most relevant work to this paper is the ParaCrawl project \cite{espla2019paracrawl}. 
It develops the Bitextor crawler and Bicleaner classifier to mining parallel data from the Internet. 
However, both components need human-labeled parallel data.
The crawler needs a labeled dictionary and the classifier needs 100K parallel sentences as the seed. 
In contrast, the proposed pipeline does not require any human-labeled data. 

There is a research line to discuss how to improve the accuracy of the parallel corpus extractor by proposing novel objective function and network architecture \cite{azpeitia2018extracting, bouamor2018h2, artetxe2018margin}. Although these methods require the supervision data to provide the training signal, we still can use the idea of this paper, generating a supervision parallel corpus in an unsupervised manner, to integrate these methods into our pipeline. 
\section{Conclusion}

In this paper, we propose an unsupervised website parallel data mining pipeline, which dose not require any labeled parallel data. The experiment results demonstrate that the machine translation systems trained with the crawled corpus are able to match the performance of the ones trained with the WMT supervision data in both rich and low resources language cases. Due to the unsupervised feature of the proposed pipeline, it can be applied to build the translation system for any language pairs that are lack of parallel corpus.
\bibliography{emnlp2020}
\bibliographystyle{acl_natbib}

\appendix

\end{document}